\documentclass{article}

\PassOptionsToPackage{numbers, compress}{natbib}

\usepackage[preprint]{neurips_2020}

\usepackage[utf8]{inputenc} 
\usepackage[T1]{fontenc}    
\usepackage{hyperref}       
\usepackage{url}            
\usepackage{booktabs}       
\usepackage{amsfonts}       
\usepackage{nicefrac}       
\usepackage{microtype}      

\usepackage{times}
\usepackage{epsfig}
\usepackage{graphicx}
\usepackage{amsmath}
\usepackage{amssymb}
\usepackage[none]{hyphenat}    
\usepackage{epstopdf}
\usepackage{subfigure}
\usepackage{bm}    
\usepackage{color} 

\usepackage[ruled,vlined]{algorithm2e}

\usepackage{multirow}                          
\usepackage{makecell}

\usepackage{authblk}

\begin{document}

\title{Neural Network Activation Quantization with Bitwise Information Bottlenecks}

%

\author[1]{Xichuan Zhou}
\author[1]{Kui Liu}
\author[1]{Cong Shi}
\author[1]{Haijun Liu}
\author[2]{Ji Liu}
\affil[1]{School of Microelectronics and Communication Engineering, Chongqing University}
\affil[2]{Ytech Seattle AI lab, FeDA lab, AI platform, Kwai Inc}

\renewcommand*{\Affilfont}{\small\it} 
\renewcommand\Authands{ and } 
\date{} 


\maketitle

\begin{abstract}
     Recent researches on information bottleneck shed new light on the continuous attempts
     to open the black box of neural signal encoding. Inspired by the problem of lossy signal compression for wireless communication,
     this paper presents a Bitwise Information Bottleneck approach for quantizing and encoding neural network activations. Based on the rate-distortion theory, the Bitwise Information
     Bottleneck attempts to determine the most significant bits in activation representation by assigning and approximating the sparse coefficient associated with each bit. Given the constraint of a limited average code
     rate, the information bottleneck minimizes the rate-distortion for optimal activation quantization in a flexible layer-by-layer manner. Experiments over
     ImageNet and other datasets show that, by minimizing the quantization rate-distortion of each layer, the neural network with information bottlenecks achieves the state-of-the-art accuracy with low-precision activation. Meanwhile, by reducing the code rate, the proposed method can improve the memory and computational efficiency by over six times compared with the deep neural network with standard single-precision representation. Codes will be available on GitHub when the paper is accepted \url{https://github.com/BitBottleneck/PublicCode}.
\end{abstract}

\section{Introduction}

    Research on activation compression for efficient neural computing is an emerging popular topic, which reveals great potential for a variety of edge-computing and computer vision applications \cite{chen2015deepdriving, wu2017squeezedet, mccool2017mixtures, xu2017end}.
    Both stochastic and deterministic approaches were proposed for neural network activation quantization \cite{gupta2015deep, courbariaux2015binaryconnect, zhou2016dorefa, wu2016quantized},
    which adopted various quantization functions to reduce the precision of activation representation
    while training the neural network. Though the loss of activation precision causes notable loss of
    classification accuracy, yet by reducing the number of bits required, the deep neural networks with quantized activation could improve both memory and time efficiency by orders of magnitude compared with the standard floating-point implementation \cite{courbariaux2015binaryconnect, kim2016bitwise, courbariaux2016binarized, rastegari2016xnor}.

    One challenge of the researches for activation quantization is the lack of theoretical ground \cite{sze2017efficient}.
    The performance of existing approaches relies on their choices of quantization strategy, and
    no minimal loss of representation precision can be determined given the average code rate.
    Recently, the concept of information bottleneck is attracting attention for its potential to
    bring a better understanding of the deep learning optimization process \cite{tishby2015deep, shwartz2017opening}. The theory of
    information bottleneck defines the relevant information in a signal $\mathbf{X}$ as being the information that this signal provides about another signal $\mathbf{\hat{X}}$. In the context of deep neural network, $\mathbf{\hat{X}}$ is
    the signal of quantized activation, and we formalize this problem as that of finding a shortcode
    for $\mathbf{\hat{X}}$ that preserves the maximum information about $\mathbf{X}$.
    That is, we minimize the loss of the information caused by signal quantization through a 'bottleneck' in the deep neural network.

    Technically, the proposed Bitwise Information Bottleneck (BIB) approach is based on the \textit{rate-distortion theory}\cite{berger2003rate}. As a lossy data compression operation, we attempt to determine the most significant bits in the activation that minimize the quantization distortion. To achieve this, the BIB approach directly minimizes the distortion of quantization given the constraint of the maximum code rate of the compressed activations. The BIB optimization is formulated as a sparse convex optimization problem, which estimates the coefficients of the most significant bits in the activation that lead to the minimal squared rate-distortion. Since the nonsignificant bits generally have near-zero coefficients, the activation can be optimally compressed in a bitwise way.

    The contributions of this paper are three-fold. First, this paper presents an information-bottleneck-based method for \textit{bitwise activation quantization} and compression. As far as we know, this is the first attempt to address the challenge of optimal activation quantization by the information-bottleneck-based method. Second, the code rates of different information bottlenecks can be tuned adaptably by a single hyperparameter of the threshold of peak-signal-to-noise-ratio (PSNR) loss, allowing the proposed method to flexibly trade-off between efficiency and accuracy for different applications.
    Finally, the Bitwise Information Bottleneck minimizes the loss of information caused by activation quantization; therefore, the proposed method suffers almost no loss of classification accuracy while obtaining over six times of memory and computational efficiency improvement.

\section{Related Work}

    Emerging research on neural encoding based on the information bottleneck is an interesting topic which attempts to answer the basic questions about the design principle of deep networks such as the optimal architecture and the optimal quantization scheme. Recently proposed deep neural networks with information bottlenecks usually had an encoder-decoder architecture for \textit{feature-wise} compression, which treated the neural network as a trade-off between compression and prediction \cite{tishby2015deep, dai2018compressing}. As far as we know, this work is the first attempt for bitwise compression addressing the challenge of minimizing the code rate while optimally preserving activation integrity.

    Despite information bottleneck motivated researches, the research on deterministic or stochastic model-based activation quantization is also attracting more attention lately. Lee quantized neural networks according to channel-level distribution \cite{lee2018quantization}. Zhao used outlier channel splitting to eliminate the effect of outliers caused by quantization \cite{zhao2019improving}. Minimum mean squared error \cite{kravchik2019low} and complementary approach \cite{banner2019post} were both used for reducing quantization error. Meanwhile, the research on data-driven activation quantization was also booming. Qiu presented a Fisher vector method to encode activation with a fixed-point number using a deep generative model \cite{qiu2017deep}. Jacob quantized both the activations and weights as 8-bit integers using an iterative approach \cite{jacob2018quantization}. Li proposed an entropy-based method for interpretable quantization \cite{li2019exploiting}. Compared with information-bottleneck-based methods, these approaches were generally based on different assumptions on the quantization function, many researchers used non-derivative quantization operation \cite{qiu2017deep, jacob2018quantization, li2019exploiting}, which might lead to uninterpretable and sub-optimal results.

    \begin{figure}[t]
    \begin{center}
    \includegraphics[width=0.7\linewidth]{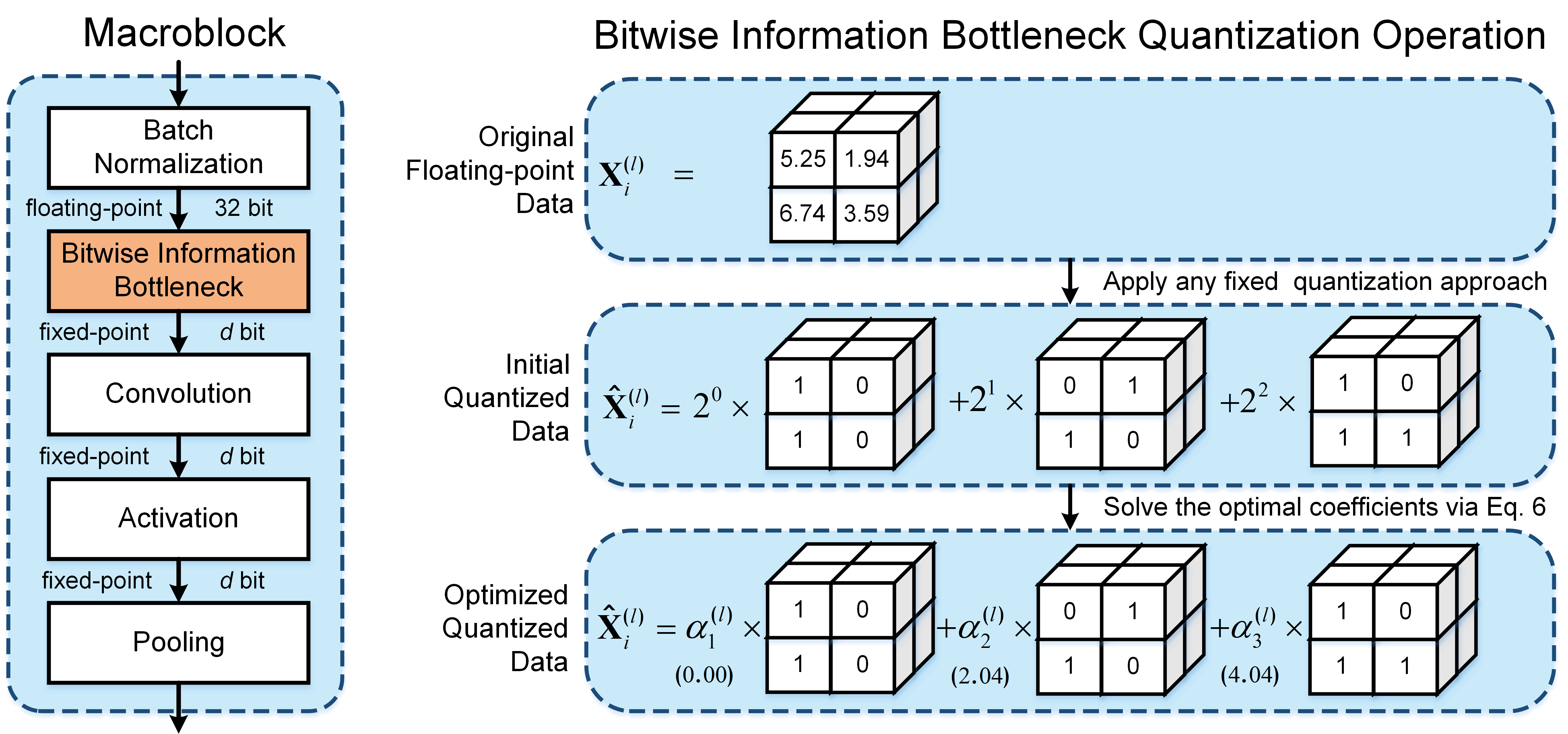}

    \end{center}
    \caption{Illustration of the Bitwise Information Bottleneck approach. The BIB operation can be inserted in the convolutional neural network for activation quantization. The idea of the proposed method is to substitute the constant levels of the standard quantization, which are the power of two in this example, for variable coefficients. By exploiting the sparsity of the bitwise coefficients, the information bottleneck could reduce the number of bits required for activation representation.}
    \label{fig:1}
    \end{figure}

\section{The Bitwise Information Bottleneck Method}

     Recently, there has been growing interest in applying information theory for deep neural network activation compression. By interpreting the deep neural network as a lossy data compression approach, the black box of the deep neural network could be opened and its performance can be optimized by the tool of rate-distortion theory, which is widely applied in the area of telecommunications\cite{shwartz2017opening}.
\subsection{Rate-distortion Theory}

     Assume $ \mathbf{{X}}_{i}^{(l)} \in \mathbb{R}^{P \cdot Q \cdot K} $ is the floating-point neural network activation tensor associated with the \textit{i}th sample output by \textit{l}th layer, where $P$, $Q$, $K$ are respectively the height, width and number of feature maps at the $l$th layer. The common quantization function $\mathcal{Q}(\cdot)$ can be written as
     \begin{equation}
      \hat{\mathbf{X}}_{i}^{(l)}=\mathcal{Q}(\mathbf{{X}}_{i}^{(l)})
      \label{CS0}
     \end{equation}

     where $\hat{\mathbf{X}}_{i}^{(l)}$ is the fixed-point representation of the floating-point activation. The goal of lossy data compression is to achieve minimal rate-distortion given the constraint of the maximum code rate. Assume $ g(\cdot) $ is the function that indicates the number of bits of the given data. According to rate-distortion theory, the typical lossy data compression approach attempts to minimize the distortion function $ d(\cdot) $ given the maximum number of bits $\eta$ over $N$ training samples as
     \begin{equation}
      \underset{\mathcal{Q}(\cdot)}{\mathrm{min}}\ \sum _{i=1}^{N}d(\mathbf{{X}}_{i}^{(l)},\mathcal{Q}(\textbf{{X}}_{i}^{(l)}))\qquad  \mathrm{s.t.}\ g(\hat{\textbf{X}}_{i}^{(l)})\leq \eta
      \label{CS1}
     \end{equation}

     In practice, the quantization function $\mathcal{Q}(\cdot)$ is nondifferentiable and nonconvex due to its integer output, which makes Eq.\ \ref{CS1} difficult to solve. Different from typical rounding-based quantization approaches that settle for suboptimal solutions, this paper attempts to find the optimal solution by reformulating Eq.\ \ref{CS1} as a sparse coding problem.

\subsection{Activation Quantization via Bitwise Bottleneck Encoding}

     Fig. \ref{fig:1} illustrates the Bitwise Information Bottleneck (BIB) approach. Formally speaking, assume $\hat{\mathbf{X}}_{i}^{(l)}$ is the $D$-bit fixed-point approximation of $\mathbf{X}_{i}^{(l)}$ as defined in Eq.\ \ref{CS2}, where $ \hat{\mathbf{X}}_{ij}^{(l)}  \in \{0, 1\}^{P \cdot Q \cdot K} $ is a three-dimensional binary tensor representing the \textit{j}th bit of $\hat{\mathbf{X}}_{i}^{(l)}$.
     \begin{equation}
     \hat{\mathbf{X}}_{i}^{(l)}=\mathcal{Q}(\mathbf{X}_{i}^{(l)})=2^{0}\hat{\mathbf{X}}_{i1}^{(l)}+2^{1}\hat{\mathbf{X}}_{i2}^{(l)}
     +...+2^{D-1}\hat{\mathbf{X}}_{iD}^{(l)}
     \label{CS2}
     \end{equation}
     where each bitwise data matrix is assigned a constant coefficient of $ {2^{0},..., 2^{D-1}} $. In practice, this bitwise activation representation allows the computation of fixed-point data to be implemented in a bitwise way. The computational and memory complexity are proportional to the number of bits of different representations.

    \begin{figure}[t]
    \begin{center}
    \subfigure[Activation distribution]{
    \includegraphics[width=0.48\linewidth]{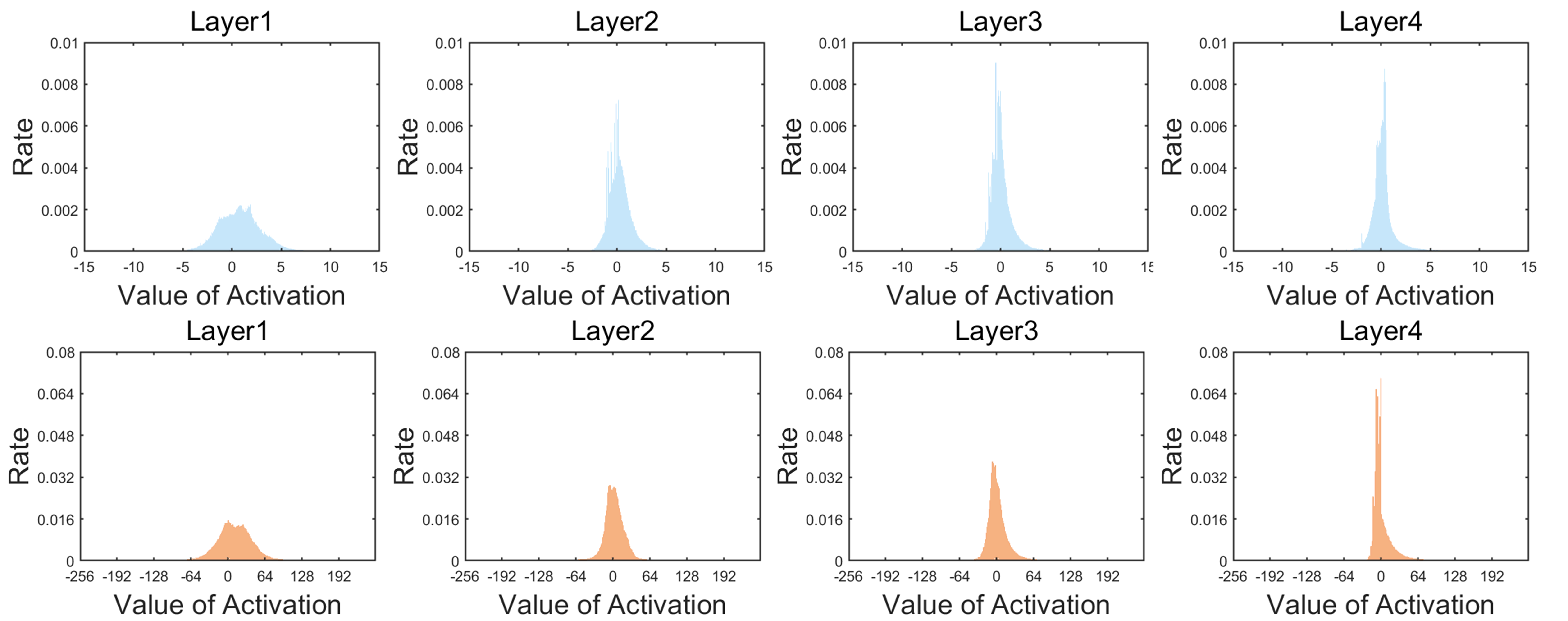}
    }
    \subfigure[Bitwise rate of code one and coefficients of $\bm{\alpha}^{(l)}$]{
    \includegraphics[width=0.48\linewidth]{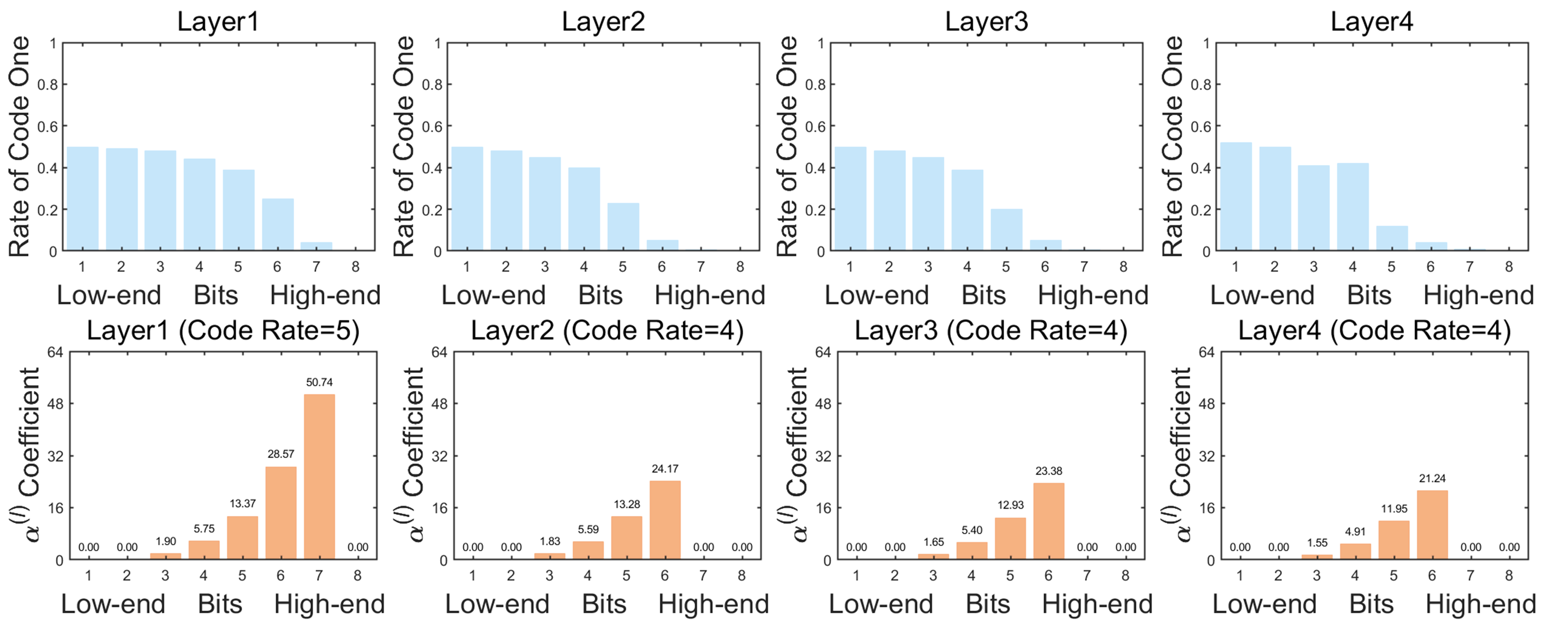}
    }%
    \end{center}
    \caption{Feature-wise vs. bitwise activation sparsity. (a) The distributions of real-valued (upper) and quantized (lower) activations of the first four layers of ResNet50 over CIFAR10. (b) The average rate of code one of each bit of the quantized activation (upper), and the estimated coefficient $\bm{\alpha}^{(l)}$ (lower). It seems that the optimal code rates of different layers depend on the level of activation sparsity, and the Bitwise Information Bottleneck can adaptively remove the near-zero high-end bits and the less-informative low-end bits of the activation representation.}
    \label{fig:2}
    \end{figure}

     Technically, the binary quantization of Eq.\ \ref{CS2} inherently assumes that each of the \textit{D} bits in the activation representation is needed, although different bits contain different but \textit{fixed} amounts of information. By removing this assumption, the proposed method substitutes the fixed coefficient for a variable $ \bm{\alpha}^{(l)} \in \mathbb{R}^{D} $as
     \begin{equation}
     \hat{\mathbf{X}}_{i}^{(l)}=\mathcal{Q}(\mathbf{X}_{i}^{(l)})=\alpha_{1}^{(l)}\hat{\mathbf{X}}_{i1}^{(l)}+\alpha
     _{2}^{(l)}\hat{\mathbf{X}}_{i2}^{(l)}+...+\alpha _{D}^{(l)}\hat{\mathbf{X}}_{iD}^{(l)}
     \label{CS3}
	 \end{equation}

     The Bitwise Information Bottleneck approach treats the design of neural networks as a trade-off between compression and prediction, which assumes that the bitwise bottlenecks can exploit the sparsity in the activation representation so that one can reduce the precision of activation representation without hurting classification accuracy. Fig.\ \ref{fig:2} illustrates the activation distributions of the first 4 layers of ResNet50 \cite{he2016deep} over the CIFAR10 dataset. It seems that activations of different layers are sparse,  but the sparsity varies from layer to layer. Fig.\ \ref{fig:2}(b) illustrates the average rate of code one in each bit of the activation representation (upper graph). Despite the change of feature-wise sparsity as shown in Fig.\ \ref{fig:2}(a), the bitwise sparsity over the high-end bits of activation representation is easier to detect. Therefore, one might be able to estimate the optimal and sparse coefficients $ \bm{\alpha}^{(l)} $ associated with the most significant bits.

     According to the rate-distortion theory (Eq.\ \ref{CS1}), the Bitwise Information Bottleneck attempts to find the optimal quantization scheme by minimizing the standard squared distortion rate over \textit{N} training samples given that less than $\eta$ bitwise coefficients are nonzero,
 	 \begin{equation}
     \underset{\bm{\alpha^{(l)}}}{\mathrm{arg}\ \mathrm{min}}\sum_{i=1}^{N} \| \mathbf{X}_{i}^{(l)}-\sum _{j=1}^{D}\alpha _{j}^{(l)}\hat{\mathbf{X}}_{ij}^{(l)} \|_{2}^{2} \quad \mathrm{s.t.}\  \| \bm{\alpha}^{(l)} \|_{0}\leq \eta
     \label{CS4}
	 \end{equation}
      where $\hat{\mathbf{X}}_{ij}^{(l)}$ calculated by initial quantization operation are usually known as the quantization \textit{codebook}. In practice, different initial quantization operations can be applied, and Eq.\ \ref{CS2} is a simple example of the rounding quantization approach. It is worth noting that since the number of nonzero coefficients in $\bm{\alpha}^{(l)}$ equals the number of bits in the fixed-point representation, the constraint function of Eq.\ \ref{CS4} actually limits the maximum number of bits in the quantized representation as required by the rate-distortion theory. Recent research shows that Eq.\ \ref{CS4} is equivalent to the following L1-norm-based problem, which leads to a sparse solution\cite{baraniuk2007compressive}.
	 \begin{equation}
     \underset{\bm{\alpha^{(l)}}}{\mathrm{arg}\ \mathrm{min}}\sum_{i=1}^{N} \| \mathbf{X}_{i}^{(l)}-\sum _{j=1}^{D}\alpha _{j}^{(l)}\hat{\mathbf{X}}_{ij}^{(l)} \|_{2}^{2} \quad \mathrm{s.t.}\ \| \bm{\alpha}^{(l)} \|_{1}\leq \eta
     \label{CS5}
	 \end{equation}
     The BIB operation solves Eq.\ \ref{CS5} to determine the sparse significant bits that lead to the minimal rate distortion. In practice, one usually
     calculates Eq. \ref{CS5} by solving its dual form as
     \begin{equation}\nonumber
     \underset{\bm{\alpha^{(l)}}}{\mathrm{arg}\ \mathrm{min}}\sum_{i=1}^{N} \| \mathbf{X}_{i}^{(l)}-\sum _{j=1}^{D}\alpha _{j}^{(l)}\hat{\mathbf{X}}_{ij}^{(l)}  \|_{2}^{2} + \lambda  \| \bm{\alpha}^{(l)} \|_{1}
     \label{CS6}
	 \end{equation}
     where $\lambda$ is the hyperparameter for controlling the trade-off between the optimized error rate and the code rate. Eq. \ref{CS5} generally leads to a sparse solution of the coefficients $ \bm{\alpha}^{(l)} $, so the activation bits associated with zero coefficients are removed during the inference stage, and the computational efficiency can be improved significantly.

\subsection{Neural Network with Bitwise Information Bottlenecks}

     As shown in Eq. \ref{CS5}, the BIB operation calculates the optimal coefficient associated with each bit of the compressed activation representation so that a minimal distortion rate can be achieved given the maximum code rate. This section shows how the Bitwise Information Bottleneck works in the deep neural network.
	
\begin{figure*}[t]
    \centering
    \includegraphics[width= 0.96\textwidth]{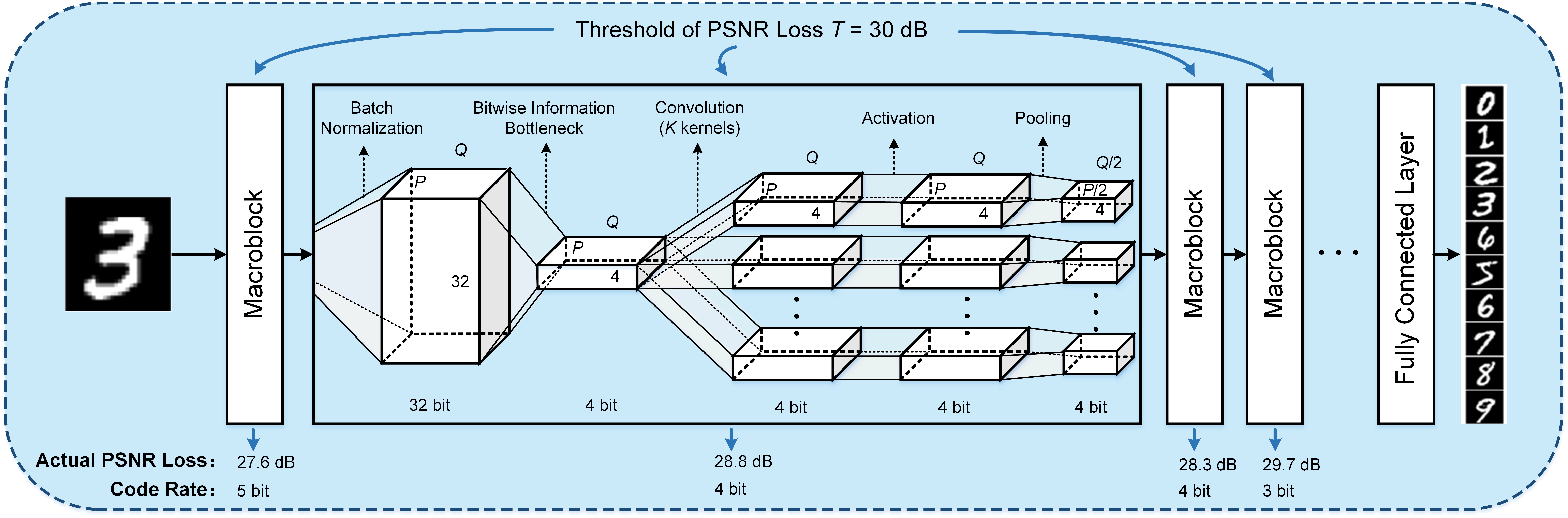}
    \caption{Deep neural network with Bitwise Information Bottleneck layers. The second macroblock is extended and shown in detail. By setting a single hyperparameter of the threshold of PSNR loss, the Bitwise Information Bottlenecks in different macroblocks can be trained to quantize the normalized activations flexibly with different optimal code rates.}
    \label{fig:3}
    \end{figure*}

     Fig.\ \ref{fig:3} shows an example of efficient neural computing with Bitwise Information Bottlenecks. The whole network is built based on the classic ResNet, although the BIB operation can be easily integrated into different networks. In practice, the BIB operation can be inserted in the macroblock of the deep neural network. A typical macroblock contains a BIB layer, a convolution layer, a pooling layer (optional), a batch-norm layer and an activation layer. Thanks to the BIB layer, which transforms the normalized floating-point activation to compressed fixed-point activation, the convolution layer can substitute the computationally expensive floating-point multiplications for efficient fixed-point bitwise multiplications.

     The benefits of inserting the Bitwise Information Bottlenecks in deep neural networks are twofold. From the perspective of \textit{memory efficiency}, compared with the standard deep neural networks using 32-bit single-precision activation representation, the Bitwise Information Bottlenecks can compress the activation into arbitrary low-precision (1 to 8 bit), obtaining an improvement of memory efficiency by 4 to 32 times. From the perspective of \textit{computational efficiency}, single-precision floating-point multiplication generally requires considerably more hardware resources and calculation time compared with fixed-point multiplication. By employing the BIB layers before the convolution layers (as shown in Fig.\ \ref{fig:1}), the inference latency of deep neural networks may be reduced by over 90$\%$, as indicated in early research\cite{zhou2018danoc}.

\subsection{Training Sparse Bitwise Information Bottlenecks}

    In practice, the sparsity level of the activation may vary from layer to layer in the neural network. It is desired that different macroblocks should use different quantization precisions. As one of the advantages of the proposed method, the optimal code rate of different Bitwise Information Bottlenecks can be approximated by tuning a single hyperparameter, which is the threshold of peak-signal-to-noise-ratio (PSNR) loss $T$. The measurement of the PSNR is defined as $\mathrm{PSNR} = 10\cdot\mathrm{log}_{10}\left ( \frac{(2^{D}-1)^2}{\mathrm{MSE}} \right )$, where $\mathrm{MSE} =\frac{1}{N\cdot P\cdot Q\cdot K}\sum_{i=1}^{N} \| \mathbf{X}_{i}^{(l)}-\sum _{j=1}^{D}\alpha _{j}^{(l)}\hat{\mathbf{X}}_{ij}^{(l)}  \|_{2}^2$. By increasing the hyperparameter of $\lambda$ and comparing the PSNR loss with the threshold, each layer can approximate the respective minimal code rate acceptable independently. More detailed information about the training algorithm can be found in Algorithm \ref{algorithm:1}.


     An illustrative example of training the neural network with multiple Bitwise Information Bottlenecks is shown in Fig.\ \ref{fig:2}. A closer look at the image reveals three interesting observations. First, the statistical phenomenon of bitwise sparsity exists. Specifically, the high-end bits of the activation representation are statistically near-zero for different layers of the neural network, which validates our assumption. Second, the optimal code rates of different layers depend on the level of activation sparsity. As an example, the activations of the first macroblock of ResNet50 are less sparse than the next three macroblocks, resulting in a higher minimal code rate acceptable. Third, it seems that the Bitwise Information Bottleneck can \textit{adaptively} reduce the coefficients $\alpha _{j}^{(l)}$ of both near-zero high-end bits and less-informative low-end bits of the activation representation.

    \begin{algorithm}[h]
    \label{algorithm:1}
    \caption{Training Algorithm}
    \LinesNumbered
    \KwIn{Pre-trained floating-point CNN model $\bm{\Theta}$,
     threshold of PSNR loss $T$,
     number of layers $L$,
     number of training samples $N$,
     number of bits for initial quantization $D$;}
    \KwOut{An Optimized Quantized model $\hat{\bm{\Theta}}$;}

    Obtain the floating-point activation tensors ${\mathbf{X}}_{i}^{(1)}, ...,{\mathbf{X}}_{i}^{(L)}$ of $\bm{\Theta}$ for each sample;

    \For{$l$ = $\mathrm{1}$,\ ...,\ $L$}
    {
        \While{$t^{(l)}$ $<$ $T$}
        {
            \For{$i$ = $\mathrm{1}$, ...,\ $N$}
            {
                Obtain the codebook $\hat{\mathbf{X}}_{i1}^{(l)}, \hat{\mathbf{X}}_{i2}^{(l)}, ..., \hat{\mathbf{X}}_{iD}^{(l)}$ by initial quantization of $\mathbf{X}_{i}^{(l)}$;

                Obtain the coefficient vector $\bm{\alpha}^{(l)}$ via Eq.\ref{CS5} and restore $\hat{\mathbf{X}}_{i}^{(l)}$ via Eq.\ref{CS3};


                Compute the PSNR loss between $\hat{\mathbf{X}}_{i}^{(l)}$ and $\mathbf{X}_{i}^{(l)}$ as $t_{i}^{(l)}$ ;

            }
            $t^{(l)}=\mathrm{max}(t_{i}^{(l)})$ for each $i = 1, ..., N$;

            Increase $\lambda$;

        }
    }
    Insert the Bitwise Information Bottleneck layers into $\bm{\Theta}$ as a new model $\hat{\bm{\Theta}}$;

    Refine and quantize the weight parameters of  $\hat{\bm{\Theta}}$ by backpropagation until reaching convergence;


    \end{algorithm}

%

\section{Experiments}



\subsection{Experimental Setting}

    Our experiments were performed on three standard benchmarks: the MNIST\cite{lecun1998gradient}, CIFAR10\cite{krizhevsky2009learning} and ImageNet (ILSVRC2012)\cite{deng2009imagenet} datasets. The MNIST dataset of $28\times28$ monochrome images contained 60 thousand training samples and 10 thousand test samples. The CIFAR10 dataset of $32\times32$ color images contained 50 thousand training samples and 10 thousand test samples, which had 10 classes. The ImageNet dataset contained 1.28 million training samples, 50 thousand verification samples, 100 thousand test samples and 1,000 classes.

    The proposed BIB approach required an initial quantization operation to calculate the binary activation codebook at the first step. Two different initial quantization methods were adopted, including the iterative rounding method \cite{courbariaux2014training} and the DoReFa method \cite{zhou2016dorefa}. To validate our method, we used the standard average classification accuracy over the test dataset as our evaluation criterion. The results of the proposed method were achieved based on the ResNet50, and all the compared approaches were also based on ResNet50 for a fair comparison.

    Similar to other studies, a backpropagation-based training process was adopted in the proposed method to refine and quantize the weight parameters of the deep neural network. In our experiment, the ADAM\cite{kingma2014adam} algorithm was applied to implement the backpropagation process, where the learning rate was 0.0001, and the batch size was 64. We trained 4,000 iterations on the MNIST and CIFAR10 datasets and 5 epochs on the ImageNet dataset.

\subsection{Visualization and Analysis}

\begin{figure}[t]
    \centering
    \subfigure[]{
    \begin{minipage}[t]{0.24\linewidth}
    \centering
    \includegraphics[width=1\linewidth]{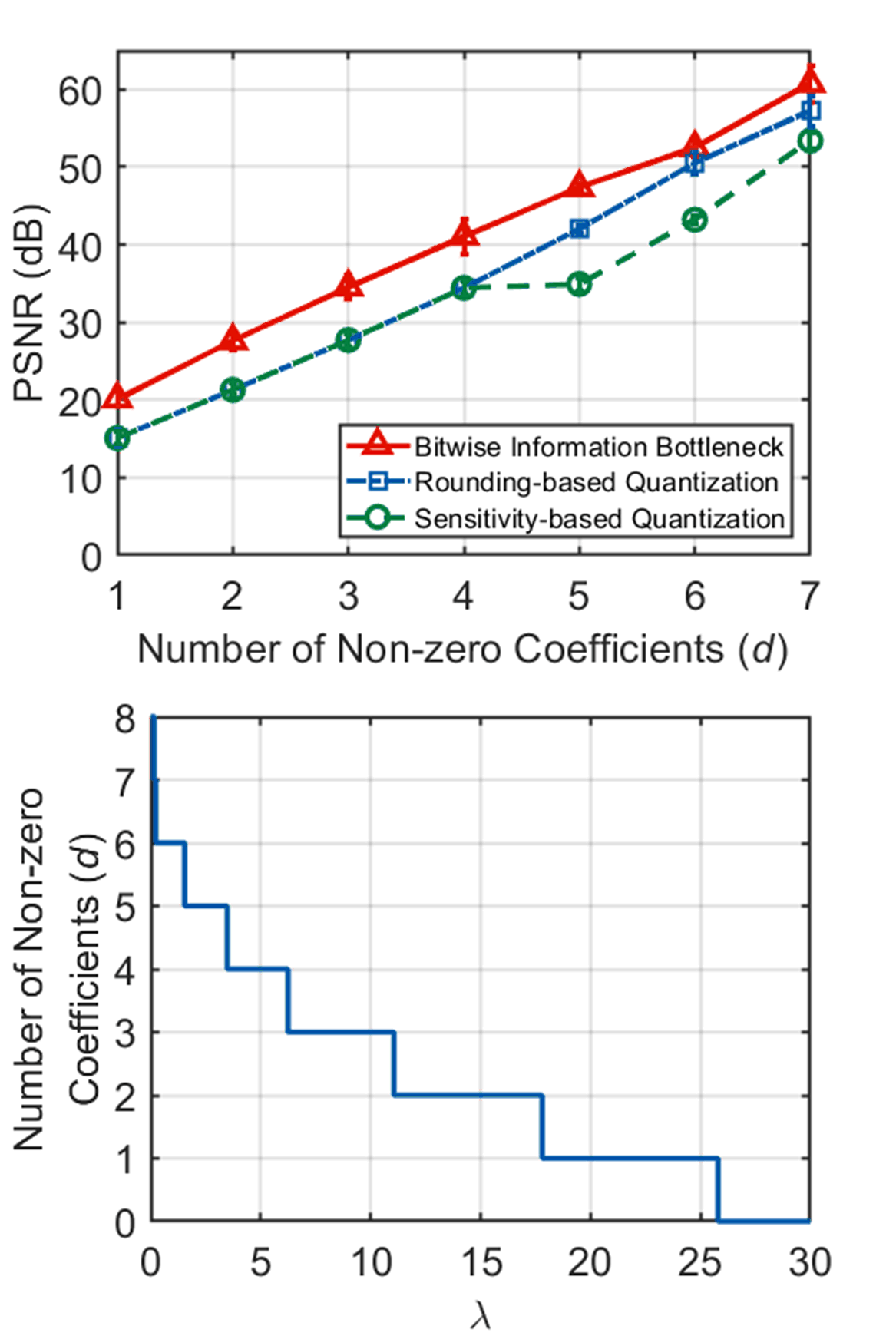}
    \end{minipage}%
    }%
    \subfigure[]{
    \begin{minipage}[t]{0.24\linewidth}
    \centering
    \includegraphics[width=1\linewidth]{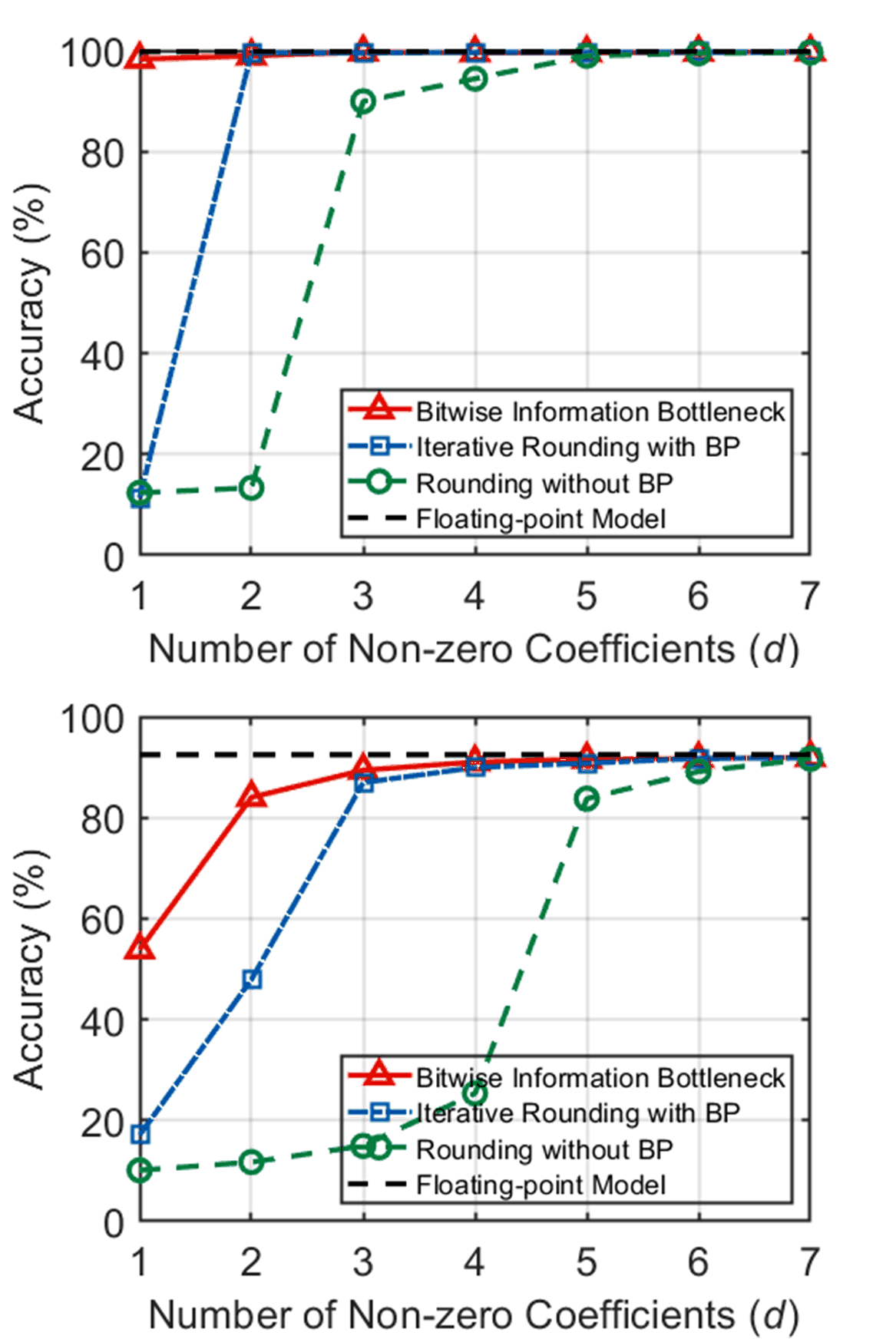}
    \end{minipage}%
    }%
    \subfigure[]{
    \begin{minipage}[t]{0.24\linewidth}
    \centering
    \includegraphics[width=1\linewidth]{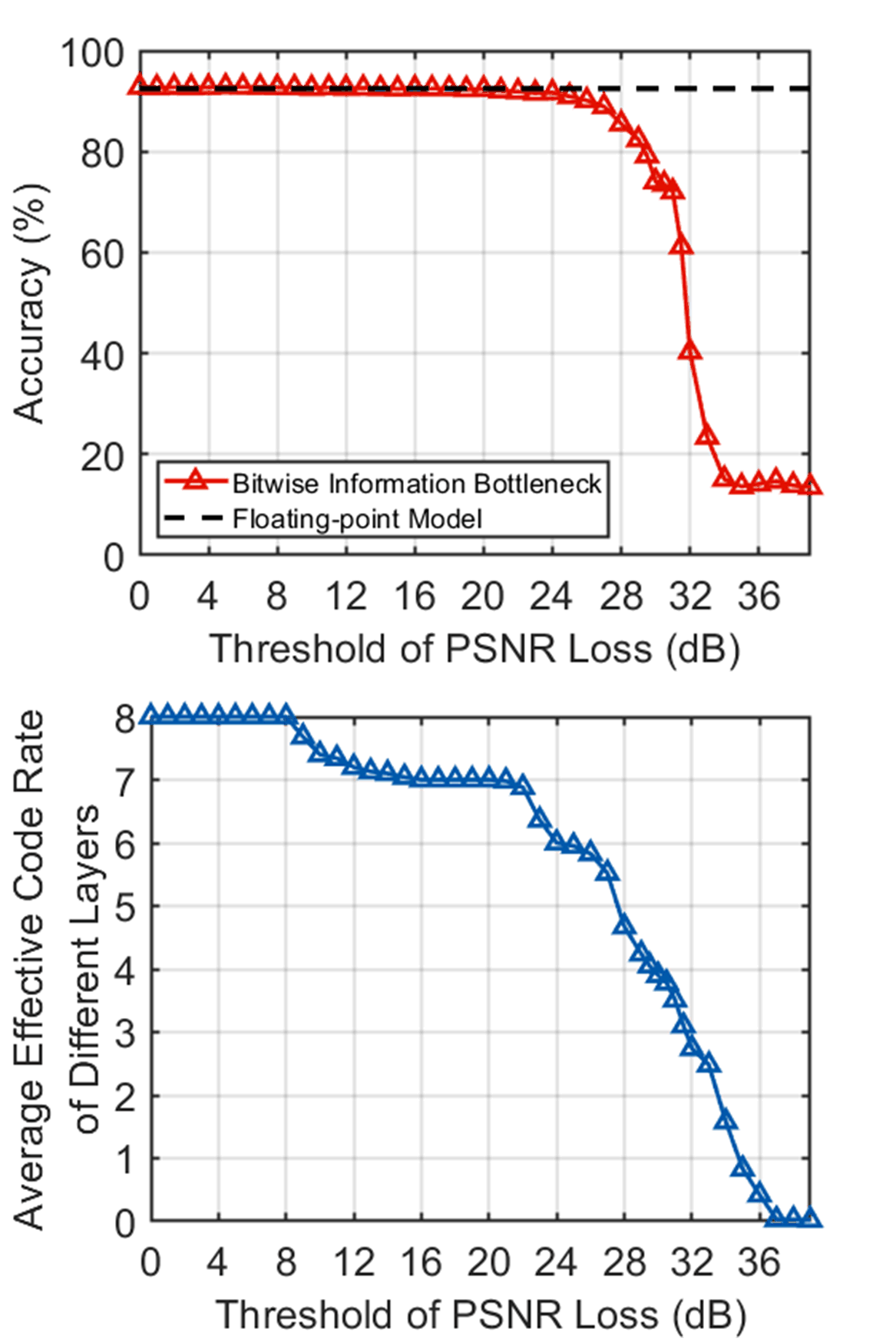}
    \end{minipage}
    }%
     \subfigure[]{
    \begin{minipage}[t]{0.24\linewidth}
    \centering
    \includegraphics[width=1\linewidth]{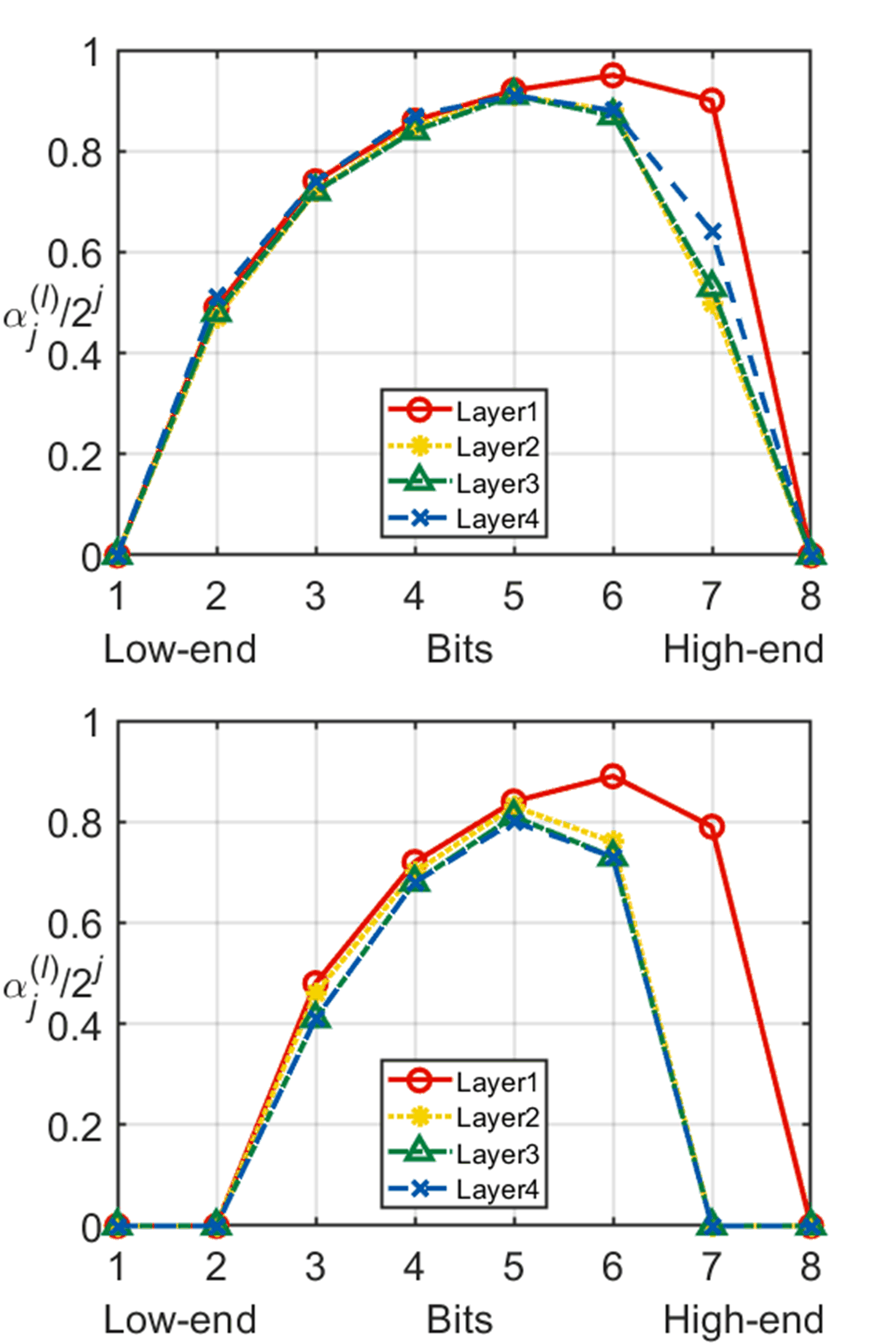}
    \end{minipage}
    }%
    \centering
    \caption{(a) The relation between the PSNR and the effective code rate, which equals the number of nonzero coefficients in $\bm{\alpha}^{(l)}$ and the relation between the effective code rate and the value of $\lambda$. (b) Accuracy of the deep neural network with Bitwise Information Bottlenecks over the MNIST (upper) and CIFAR10 (lower) datasets, which can compress the activation of the deep neural network with different effective code rates \emph{d}. (c) The change in accuracy and average effective code rate of different layers when using the Bitwise Information Bottleneck to compress the 8-bit fixed-point activation according to different PSNR loss thresholds. (d) The ratio of $\alpha^{(l)}_{j}/2^{j}$ with different threshold of PSNR loss ($T=26$ upper, $T=30$ lower).}
    \label{fig:4}
    \end{figure}

    A group of experiments was performed over the MNIST dataset to analyze and visualize the performance of the proposed method compared with the initial and other quantization approaches. The Bitwise Information Bottleneck reduced the code rate of the fixed-point representation from the initial quantization length of $\textit{D}$ bits to the effective code rate $\textit{d}$, which was the number of nonzero coefficients in vector $\bm{\alpha}^{(l)}$ (Eq. \ref{CS5}). Fig. \ref{fig:4}(a) shows the relation between the effective code rate $d$ and the average PSNR achieved over the MNIST dataset. Overall, the proposed BIB approach achieved 1.98 dB to 12.44 dB higher PSNR compared with the classic rounding-based quantization and sensitivity-based quantization methods \cite{li2020build}. The superiority of PSNR was consistent when different numbers of nonzero coefficients were chosen, which can be controlled by the hyperparameter of $\lambda$, as shown in the figure.

    Similar to other activation quantization approaches, a backpropagation retraining process was applied after the activation
    quantization process to refine the weight parameters in the proposed method. Fig. \ref{fig:4}(b) shows the final classification accuracy achieved with backpropagation weight refinement. The average accuracy over both the MNIST and CIFAR10 datasets was evaluated with various numbers of nonzero coefficients $d$.
    Experimental results showed that the proposed method outperformed the baseline
    approach when fewer than four bits were used for activation quantization. The loss of accuracy caused by the rounding-based quantization became worse when the dataset was more complicated, while the BIB approach suffered almost
    no loss of accuracy compared with the floating-point model when more than three bits were used for activation representation.

    The BIB approach featured its ability to automatically determine the effective code rates of different
    layers. By setting the threshold of PSNR loss accepted for activation quantization, the Bitwise Information Bottleneck could estimate
    the minimal effective code rate by sparse optimization. Fig. \ref{fig:4}(c) shows the relation among the threshold
    of PSNR loss, the average effective code rate across different layers, and the classification accuracy.
    As shown in our experiments over the CIFAR10 dataset, the classification accuracy declined when the threshold of PSNR loss increased.
    However, when the PSNR loss was less than 24 dB, almost no decrease in classification accuracy was detected (less than 1\%). The decrease in accuracy began when fewer than 6 bits on average were used for activation representation.

    The Bitwise Information Bottleneck handles the task of activation quantization as a trade-off between compression and prediction. Fig.\ \ref{fig:4}(d) shows the ratio of $\alpha^{(l)}_{j}/2^{j}$ for each bit of the quantized activation, where $2^{j}$ is the natural coefficient of Eq.\ \ref{CS2}, which reflects how much information of the $j$th bit contains. Two interesting observations can be found in the image. First, the low-end bits of the activation representation were removed by the information bottleneck, leading to minimal loss of information caused by bit compression. Second, despite containing more information, the high-end bits were also removed by the information bottleneck, which was caused by the bitwise sparsity in high-end bits (as shown in Fig.\ \ref{fig:2}).

\subsection{Compared with the State-of-the-Art}

    We evaluated the proposed method with the state-of-the-art activation quantization approaches over the ImageNet dataset.
    The original floating-point model and the proposed BIB approach were compared with the ACIQ \cite{banner2019post}, Focused compression \cite{zhao2019focused}, Integer-only \cite{jacob2018quantization}, UNIQ \cite{baskin2018uniq}, INQ\cite{zhou2017incremental}, DoReFa-Net\cite{zhou2016dorefa} and Iterative rounding \cite{courbariaux2014training} approaches. A threshold of PSNR loss of 16 dB and 8 dB was adopted to quantize the activation of the deep neural network, which resulted in an average effective code rate of $4.0(\pm1)$ and $5.0(\pm1)$ bits. The quantization method of DoReFa-Net\cite{zhou2016dorefa} was adopted for initial quantization.
    The standard TensorFlow quantization tool was used to quantize the weight to 8 bits.

     Table.\ \ref{Table:1} summarizes the results of classification accuracies over the ImageNet dataset.
     In summary, the proposed method achieved the highest classification accuracy when five bits were adopted
     for activation representation. The BIB approach with 5-bit quantization suffered almost no loss of
     accuracy (less than 1\%) compared with the floating-point model.
     The BIB approach with 5-bit activation representation achieved similar accuracy with the Integer-only approach with 8-bit activation and achieved similar results with the INQ and Focused compression approach with 32-bit floating-point activation.
     When using 4-bit fixed-point representation, the performance of the BIB approach was
     still competitive with the state-of-the-art.

    \begin{table*}[t]
    \centering
    \caption{Comparing with state-of-the-art over the ImageNet dataset.}
    \resizebox{0.8\linewidth}{!}{
	\begin{tabular}{l|c|cc|cc}
		\hline
        \hline
         \textbf{Method}               & \textbf{Year}               & \textbf{Weights}      &  \textbf{Activations}       &\textbf{Top-1 Acc.}       &\textbf{Top-5 Acc.}          \\
        \hline
          Floating-point Model         &2016       &32            &32                   &75.6$\%$            &92.8$\%$                \\
          \hline
          \textbf{Bitwise Information Bottleneck} &--      &8             &5                    &\textbf{75.1}$\textbf{\%}$           &\textbf{92.5}$\textbf{\%}$               \\
          Focused compression    & 2019        &5             &32                    &74.9\%             &92.6\%                      \\
          Integer-only           & 2018      &8             &8                    &74.9$\%$            &--                     \\
          INQ                    & 2017        &5             &32                   &74.8$\%$            &92.5$\%$                  \\
		  DoReFa-Net (initial quantization)  & 2016              &8             &5                    &73.8$\%$            &91.7$\%$               \\
          Iterative rounding     & 2014          &8            &5                    &72.1\%              &90.4\%                         \\
           \hline
          \textbf{Bitwise Information Bottleneck}   &--          &8             &4                    &\textbf{73.7}$\textbf{\%}$           &\textbf{91.7}$\textbf{\%}$                            \\
          UNIQ                      & 2018        &4             &8                    &73.4\%              &--                                      \\
          ACIQ                      &2019         &8             &4                    &71.8\%              &--                                     \\
          DoReFa-Net (initial quantization)    & 2016            &8             &4                    &70.1$\%$            &89.3$\%$                              \\
          Iterative rounding       & 2014    &8             &4                    &70.0$\%$            &89.4$\%$                              \\

		  \hline

	\end{tabular}
    }
    \label{Table:1}
    \end{table*}

    \begin{table*}[t]
    \centering
    \caption{The memory and computational efficiency improvement achieved by the BIB operations with different effective code rates. In the table, M indicates megabytes, B indicates billion.}
    \resizebox{0.8\linewidth}{!}{
        \begin{tabular}{l|rrrrrrrrr}
		\hline\hline
        \textbf{Number of Bits}          &\textbf{1 bit}            &\textbf{2 bit}            &\textbf{3 bit}            &\textbf{4 bit}            &\textbf{5 bit}            &\textbf{6 bit}            &\textbf{7 bit}            &\textbf{8 bit}            &\textbf{32 bit}           \\
        \hline
		Operation      &8.9B        &17.8B        &26.7B        &35.6B        &44.5B        &53.4B        &62.3B        &71.3B        &285.0B \\

		Memory        &1.1M         &2.1M         &3.2M         &4.3M         &5.3M         &6.4M         &7.4M         &8.5M         &34.0M         \\

        Improvement  &30.9$\times$  &16.2$\times$  &10.6$\times$  &7.9$\times$   &6.4$\times$  &5.3$\times$  &4.6$\times$  &4.0$\times$  &1.0$\times$   \\
		\hline
	    \end{tabular}
        }

        \label{Table:2}
    \end{table*}

\subsection{Efficiency Improvement}
    The BIB approach attempts to reduce the complexity of the inference stage for embedded computer vision applications. As shown in Table\ \ref{Table:2}, compared with the 32-bit activation representation, the Bitwise Information Bottleneck could improve the computational efficiency by over 6.4 times without hurting the performance. From the perspective of reducing memory occupation, the running memory can be reduced by up to 84\%.

%

\section{Conclusion and Discussion}
    This paper presents a Bitwise Information Bottleneck approach for neural network activation quantization. Based on the rate-distortion theory, the Bitwise Information Bottlenecks inserted in the deep neural network attempt to minimize the quantization error rate and code rate. Experiments over different datasets show that the proposed method could reduce the activation of the deep neural network to one to five bits without hurting the performance, which could improve the memory efficiency as well as the computational efficiency of the neural-network inference by over 6 times.

    The proposed Bitwise Information Bottleneck method assumes that the activation of deep neural networks has a certain level of sparsity; however, in practice, the activation of different layers has a different level of sparsity. Although applying flexible quantization as adopted by different Bitwise Information Bottlenecks could reduce the quantization PSNR loss, the proposed method is subject to the constraint of activation sparsity. In the future, we plan to incorporate sparse regularization to intentionally introduce sparsity in the activation, so that the Bitwise Information Bottlenecks could achieve higher activation integrity with a lower code rate.

{\small
\bibliographystyle{plainnat}
\bibliography{bibfile}

\begin{thebibliography}{35}
\providecommand{\natexlab}[1]{#1}
\providecommand{\url}[1]{\texttt{#1}}
\expandafter\ifx\csname urlstyle\endcsname\relax
  \providecommand{\doi}[1]{doi: #1}\else
  \providecommand{\doi}{doi: \begingroup \urlstyle{rm}\Url}\fi

\bibitem[Banner et~al.(2019)Banner, Nahshan, and Soudry]{banner2019post}
Ron Banner, Yury Nahshan, and Daniel Soudry.
\newblock Post training 4-bit quantization of convolutional networks for
  rapid-deployment.
\newblock In \emph{Advances in Neural Information Processing Systems}, pages
  7948--7956, 2019.

\bibitem[Baraniuk(2007)]{baraniuk2007compressive}
Richard~G Baraniuk.
\newblock Compressive sensing.
\newblock \emph{IEEE signal processing magazine}, 24\penalty0 (4), 2007.

\bibitem[Baskin et~al.(2018)Baskin, Schwartz, Zheltonozhskii, Liss, Giryes,
  Bronstein, and Mendelson]{baskin2018uniq}
Chaim Baskin, Eli Schwartz, Evgenii Zheltonozhskii, Natan Liss, Raja Giryes,
  Alex~M Bronstein, and Avi Mendelson.
\newblock Uniq: uniform noise injection for the quantization of neural
  networks.
\newblock \emph{arXiv preprint arXiv:1804.10969}, 3, 2018.

\bibitem[Berger(2003)]{berger2003rate}
Toby Berger.
\newblock Rate-distortion theory.
\newblock \emph{Wiley Encyclopedia of Telecommunications}, 2003.

\bibitem[Chen et~al.(2015)Chen, Seff, Kornhauser, and
  Xiao]{chen2015deepdriving}
Chenyi Chen, Ari Seff, Alain Kornhauser, and Jianxiong Xiao.
\newblock Deepdriving: Learning affordance for direct perception in autonomous
  driving.
\newblock In \emph{Proceedings of the IEEE International Conference on Computer
  Vision}, pages 2722--2730, 2015.

\bibitem[Courbariaux et~al.(2014)Courbariaux, Bengio, and
  David]{courbariaux2014training}
Matthieu Courbariaux, Yoshua Bengio, and Jean-Pierre David.
\newblock Training deep neural networks with low precision multiplications.
\newblock \emph{arXiv preprint arXiv:1412.7024}, 2014.

\bibitem[Courbariaux et~al.(2015)Courbariaux, Bengio, and
  David]{courbariaux2015binaryconnect}
Matthieu Courbariaux, Yoshua Bengio, and Jean-Pierre David.
\newblock Binaryconnect: Training deep neural networks with binary weights
  during propagations.
\newblock In \emph{Advances in neural information processing systems}, pages
  3123--3131, 2015.

\bibitem[Courbariaux et~al.(2016)Courbariaux, Hubara, Soudry, El-Yaniv, and
  Bengio]{courbariaux2016binarized}
Matthieu Courbariaux, Itay Hubara, Daniel Soudry, Ran El-Yaniv, and Yoshua
  Bengio.
\newblock Binarized neural networks: Training deep neural networks with weights
  and activations constrained to+ 1 or-1.
\newblock \emph{arXiv preprint arXiv:1602.02830}, 2016.

\bibitem[Dai et~al.(2018)Dai, Zhu, and Wipf]{dai2018compressing}
Bin Dai, Chen Zhu, and David Wipf.
\newblock Compressing neural networks using the variational information
  bottleneck.
\newblock \emph{arXiv preprint arXiv:1802.10399}, 2018.

\bibitem[Deng et~al.(2009)Deng, Dong, Socher, Li, Li, and
  Fei-Fei]{deng2009imagenet}
Jia Deng, Wei Dong, Richard Socher, Li-Jia Li, Kai Li, and Li~Fei-Fei.
\newblock Imagenet: A large-scale hierarchical image database.
\newblock In \emph{2009 IEEE conference on computer vision and pattern
  recognition}, pages 248--255. Ieee, 2009.

\bibitem[Gupta et~al.(2015)Gupta, Agrawal, Gopalakrishnan, and
  Narayanan]{gupta2015deep}
Suyog Gupta, Ankur Agrawal, Kailash Gopalakrishnan, and Pritish Narayanan.
\newblock Deep learning with limited numerical precision.
\newblock In \emph{International Conference on Machine Learning}, pages
  1737--1746, 2015.

\bibitem[He et~al.(2016)He, Zhang, Ren, and Sun]{he2016deep}
Kaiming He, Xiangyu Zhang, Shaoqing Ren, and Jian Sun.
\newblock Deep residual learning for image recognition.
\newblock In \emph{Proceedings of the IEEE conference on computer vision and
  pattern recognition}, pages 770--778, 2016.

\bibitem[Jacob et~al.(2018)Jacob, Kligys, Chen, Zhu, Tang, Howard, Adam, and
  Kalenichenko]{jacob2018quantization}
Benoit Jacob, Skirmantas Kligys, Bo~Chen, Menglong Zhu, Matthew Tang, Andrew
  Howard, Hartwig Adam, and Dmitry Kalenichenko.
\newblock Quantization and training of neural networks for efficient
  integer-arithmetic-only inference.
\newblock In \emph{Proceedings of the IEEE Conference on Computer Vision and
  Pattern Recognition}, pages 2704--2713, 2018.

\bibitem[Kim and Smaragdis(2016)]{kim2016bitwise}
Minje Kim and Paris Smaragdis.
\newblock Bitwise neural networks.
\newblock \emph{arXiv preprint arXiv:1601.06071}, 2016.

\bibitem[Kingma and Ba(2014)]{kingma2014adam}
Diederik~P Kingma and Jimmy Ba.
\newblock Adam: A method for stochastic optimization.
\newblock \emph{arXiv preprint arXiv:1412.6980}, 2014.

\bibitem[Kravchik et~al.(2019)Kravchik, Yang, Kisilev, and
  Choukroun]{kravchik2019low}
Eli Kravchik, Fan Yang, Pavel Kisilev, and Yoni Choukroun.
\newblock Low-bit quantization of neural networks for efficient inference.
\newblock In \emph{Proceedings of the IEEE International Conference on Computer
  Vision Workshops}, pages 0--0, 2019.

\bibitem[Krizhevsky et~al.(2009)Krizhevsky, Hinton,
  et~al.]{krizhevsky2009learning}
Alex Krizhevsky, Geoffrey Hinton, et~al.
\newblock Learning multiple layers of features from tiny images.
\newblock Technical report, Citeseer, 2009.

\bibitem[LeCun et~al.(1998)LeCun, Bottou, Bengio, Haffner,
  et~al.]{lecun1998gradient}
Yann LeCun, L{\'e}on Bottou, Yoshua Bengio, Patrick Haffner, et~al.
\newblock Gradient-based learning applied to document recognition.
\newblock \emph{Proceedings of the IEEE}, 86\penalty0 (11):\penalty0
  2278--2324, 1998.

\bibitem[Lee et~al.(2018)Lee, Ha, Choi, Lee, and Lee]{lee2018quantization}
Jun~Haeng Lee, Sangwon Ha, Saerom Choi, Won-Jo Lee, and Seungwon Lee.
\newblock Quantization for rapid deployment of deep neural networks.
\newblock \emph{arXiv preprint arXiv:1810.05488}, 2018.

\bibitem[Li et~al.(2020)Li, Zhang, Zhou, and Ren]{li2020build}
Yixing Li, Shuai Zhang, Xichuan Zhou, and Fengbo Ren.
\newblock Build a compact binary neural network through bit-level sensitivity
  and data pruning.
\newblock \emph{Neurocomputing}, 2020.

\bibitem[Li et~al.(2019)Li, Lin, Zhang, Liu, Doermann, Wu, Huang, and
  Ji]{li2019exploiting}
Yuchao Li, Shaohui Lin, Baochang Zhang, Jianzhuang Liu, David Doermann,
  Yongjian Wu, Feiyue Huang, and Rongrong Ji.
\newblock Exploiting kernel sparsity and entropy for interpretable cnn
  compression.
\newblock In \emph{Proceedings of the IEEE Conference on Computer Vision and
  Pattern Recognition}, pages 2800--2809, 2019.

\bibitem[McCool et~al.(2017)McCool, Perez, and Upcroft]{mccool2017mixtures}
Chris McCool, Tristan Perez, and Ben Upcroft.
\newblock Mixtures of lightweight deep convolutional neural networks: Applied
  to agricultural robotics.
\newblock \emph{IEEE Robotics and Automation Letters}, 2\penalty0 (3):\penalty0
  1344--1351, 2017.

\bibitem[Qiu et~al.(2017)Qiu, Yao, and Mei]{qiu2017deep}
Zhaofan Qiu, Ting Yao, and Tao Mei.
\newblock Deep quantization: Encoding convolutional activations with deep
  generative model.
\newblock In \emph{Proceedings of the IEEE Conference on Computer Vision and
  Pattern Recognition}, pages 6759--6768, 2017.

\bibitem[Rastegari et~al.(2016)Rastegari, Ordonez, Redmon, and
  Farhadi]{rastegari2016xnor}
Mohammad Rastegari, Vicente Ordonez, Joseph Redmon, and Ali Farhadi.
\newblock Xnor-net: Imagenet classification using binary convolutional neural
  networks.
\newblock In \emph{European Conference on Computer Vision}, pages 525--542.
  Springer, 2016.

\bibitem[Shwartz-Ziv and Tishby(2017)]{shwartz2017opening}
Ravid Shwartz-Ziv and Naftali Tishby.
\newblock Opening the black box of deep neural networks via information.
\newblock \emph{arXiv preprint arXiv:1703.00810}, 2017.

\bibitem[Sze et~al.(2017)Sze, Chen, Yang, and Emer]{sze2017efficient}
Vivienne Sze, Yu-Hsin Chen, Tien-Ju Yang, and Joel~S Emer.
\newblock Efficient processing of deep neural networks: A tutorial and survey.
\newblock \emph{Proceedings of the IEEE}, 105\penalty0 (12):\penalty0
  2295--2329, 2017.

\bibitem[Tishby and Zaslavsky(2015)]{tishby2015deep}
Naftali Tishby and Noga Zaslavsky.
\newblock Deep learning and the information bottleneck principle.
\newblock In \emph{2015 IEEE Information Theory Workshop (ITW)}, pages 1--5.
  IEEE, 2015.

\bibitem[Wu et~al.(2017)Wu, Iandola, Jin, and Keutzer]{wu2017squeezedet}
Bichen Wu, Forrest Iandola, Peter~H Jin, and Kurt Keutzer.
\newblock Squeezedet: Unified, small, low power fully convolutional neural
  networks for real-time object detection for autonomous driving.
\newblock In \emph{Proceedings of the IEEE Conference on Computer Vision and
  Pattern Recognition Workshops}, pages 129--137, 2017.

\bibitem[Wu et~al.(2016)Wu, Leng, Wang, Hu, and Cheng]{wu2016quantized}
Jiaxiang Wu, Cong Leng, Yuhang Wang, Qinghao Hu, and Jian Cheng.
\newblock Quantized convolutional neural networks for mobile devices.
\newblock In \emph{Proceedings of the IEEE Conference on Computer Vision and
  Pattern Recognition}, pages 4820--4828, 2016.

\bibitem[Xu et~al.(2017)Xu, Gao, Yu, and Darrell]{xu2017end}
Huazhe Xu, Yang Gao, Fisher Yu, and Trevor Darrell.
\newblock End-to-end learning of driving models from large-scale video
  datasets.
\newblock In \emph{Proceedings of the IEEE conference on computer vision and
  pattern recognition}, pages 2174--2182, 2017.

\bibitem[Zhao et~al.(2019{\natexlab{a}})Zhao, Hu, Dotzel, De~Sa, and
  Zhang]{zhao2019improving}
Ritchie Zhao, Yuwei Hu, Jordan Dotzel, Christopher De~Sa, and Zhiru Zhang.
\newblock Improving neural network quantization using outlier channel
  splitting.
\newblock \emph{arXiv preprint arXiv:1901.09504}, 2019{\natexlab{a}}.

\bibitem[Zhao et~al.(2019{\natexlab{b}})Zhao, Gao, Bates, Mullins, and
  Xu]{zhao2019focused}
Yiren Zhao, Xitong Gao, Daniel Bates, Robert Mullins, and Cheng-Zhong Xu.
\newblock Focused quantization for sparse cnns.
\newblock In \emph{Advances in Neural Information Processing Systems}, pages
  5585--5594, 2019{\natexlab{b}}.

\bibitem[Zhou et~al.(2017)Zhou, Yao, Guo, Xu, and Chen]{zhou2017incremental}
Aojun Zhou, Anbang Yao, Yiwen Guo, Lin Xu, and Yurong Chen.
\newblock Incremental network quantization: Towards lossless cnns with
  low-precision weights.
\newblock \emph{arXiv preprint arXiv:1702.03044}, 2017.

\bibitem[Zhou et~al.(2016)Zhou, Wu, Ni, Zhou, Wen, and Zou]{zhou2016dorefa}
Shuchang Zhou, Yuxin Wu, Zekun Ni, Xinyu Zhou, He~Wen, and Yuheng Zou.
\newblock Dorefa-net: Training low bitwidth convolutional neural networks with
  low bitwidth gradients.
\newblock \emph{arXiv preprint arXiv:1606.06160}, 2016.

\bibitem[Zhou et~al.(2018)Zhou, Li, Tang, Hu, Lin, and Zhang]{zhou2018danoc}
X.~Zhou, S.~Li, F.~Tang, S.~Hu, Z.~Lin, and L.~Zhang.
\newblock Danoc: An efficient algorithm and hardware codesign of deep neural
  networks on chip.
\newblock \emph{IEEE Trans Neural Netw Learn Syst}, 29\penalty0 (7):\penalty0
  3176--3187, 2018.

\end{thebibliography}
}

%
%
%
%
%

\end{document}